# Capturing knowledge of user preferences: ontologies in recommender systems

**Stuart E. Middleton, David C. De Roure and Nigel R. Shadbolt**
Department of Electronics and Computer Science
University of Southampton
Southampton, S017 1BJ, UK
Email : {sem99r,dder,nrs}@ecs.soton.ac.uk

**ABSTRACT**
Tools for filtering the World Wide Web exist, but they are hampered by the difficulty of capturing user preferences in such a dynamic environment. We explore the acquisition of user profiles by unobtrusive monitoring of browsing behaviour and application of supervised machine-learning techniques coupled with an ontological representation to extract user preferences. A multi-class approach to paper classification is used, allowing the paper topic taxonomy to be utilised during profile construction. The Quickstep recommender system is presented and two empirical studies evaluate it in a real work setting, measuring the effectiveness of using a hierarchical topic ontology compared with an extendable flat list.

## INTRODUCTION

The mass of content available on the World-Wide Web raises important questions over its effective use. With largely unstructured pages authored by a massive range of people on a diverse range of topics, simple browsing has given way to filtering as the practical way to manage web-based information – and for most of us that means search engines.

Search engines are very effective at filtering pages that match explicit queries. Unfortunately, most people find articulating what they want extremely difficult, especially if forced to use a limited vocabulary such as keywords. The result is large lists of search results that contain a handful of useful pages, defeating the purpose of filtering in the first place.

### Recommender Systems Can Help

Now people may find articulating what they want hard, but they are very good at recognizing it when they see it. This insight has led to the utilization of relevance feedback, where people rate web pages as interesting or not interesting and the system tries to find pages that match the interesting examples (positive examples) and do not match the not interesting examples (negative examples). With sufficient positive and negative examples, modern machine learning techniques can classify new pages with impressive accuracy.

Obtaining sufficient examples is difficult however, especially when trying to obtain negative examples. The problem with asking people for examples is that the cost, in terms of time and effort, of providing the examples generally outweighs the reward they will eventually receive. Negative examples are particularly unrewarding, since there could be many irrelevant items to any typical query.

Unobtrusive monitoring provides positive examples of what the user is looking for, without interfering with the users normal activity. Heuristics can also be applied to infer negative examples, although generally with less confidence. This idea has led to content-based recommender systems, which unobtrusively watch users browse the web, and recommend new pages that correlate with a user profile.

Another way to recommend pages is based on the ratings of other people who have seen the page before. Collaborative recommender systems do this by asking people to rate explicitly pages and then recommend new pages that similar users have rated highly. The problem with collaborative filtering is that there is no direct reward for providing examples since they only help other people. This leads to initial difficulties in obtaining a sufficient number of ratings for the system to be useful.

Hybrid systems, attempting to combine the advantages of content-based and collaborative recommender systems, have proved popular to-date. The feedback required for content-based recommendation is shared, allowing collaborative recommendation as well. A hybrid approach is used by our Quickstep recommender system.

This work follows the tradition of over 30 years of knowledge acquisition. Knowledge acquisition above the normal workflow is intrusive and counterproductive. We present a system with a low level of intrusiveness, driven by people making explicit choices that reflect the real world to capture profiles.

### The Problem Domain

As the trend to publish research papers on-line increases, researchers are increasingly using the web as their primary source of papers. Typical researchers need to know about new papers in their general field of interest, and older papers relating to their current work. In addition, researchers time is limited, as browsing competes with other tasks in the work place. It is this problem our Quickstep recommender system addresses.

Since researchers have their usual work to perform, unobtrusive monitoring methods are preferred else they will be reluctant to use the system. Also, very high recommendation accuracy is not critical as long as the system is deemed useful to them.

Evaluation of real world knowledge acquisition systems, as Shadbolt [21] discusses, is both tricky and complex. A lot of evaluations are performed with user log data (simulating real user activity) or with standard benchmark collections. Although these evaluations are useful, especially for technique comparison, they must be backed up by real world studies so we can see how the benchmark tests generalize to the real world setting. Similar problems are seen in the agent domain where, as Nwana [16] argues, it has yet to be conclusively demonstrated if people really benefit from such information systems.

This is why we have chosen a real problem upon which to evaluate our Quickstep recommender system.

### User Profiling in Recommender Systems

User modelling is typically either knowledge-based or behaviour-based. Knowledge-based approaches engineer static models of users and dynamically match users to the closest model. Behaviour-based approaches use the users behaviour itself as a model, often using machine-learning techniques to discover useful patterns of behaviour. Kobsa [10] provides a good survey of user modelling techniques.

The typical user profiling approach for recommender systems is behaviour-based, using a binary model representing what users find interesting and uninteresting. Machine-learning techniques are then used to assess potential items of interest in respect to the binary model. There are a lot of effective machine learning algorithms based on two classes. Sebastiani [20] provides a good survey of current machine learning techniques and De Roure [5] a review of recommender systems.

Although more difficult than the binary case, we choose to use a multi-class behavioural model. This allows the classes to represent paper topics, and hence domain knowledge to be used when constructing the user profile. We thus bring together ideas from knowledge-based and behaviour-based modelling to address the problem domain.

### Ontology Use and the World Wide Web

Ontologies are used both to structure the web, as in Yahoo's search space categorization, and to provide a common basis for understanding between systems, such as in the knowledge query modelling language (KQML). In-depth ontological representations are also seen in knowledge-based systems, which use relationships between web entities (bookmarks, web pages, page authors etc.) to infer facts about given situations.

We use an ontology to investigate how domain knowledge can help in the acquisition of user preferences.

### Overview of the Quickstep System

Quickstep unobtrusively monitors user browsing behaviour via a proxy server, logging each URL browsed during normal work activity. A machine-learning algorithm classifies browsed URLs overnight, and saves each classified paper in a central paper store. Explicit feedback and browsed topics form the basis of the interest profile for each user.

Each day a set of recommendations is computed, based on correlations between user interest profiles and classified paper topics. Any feedback offered on these recommendations is recorded when the user looks at them.

Users can provide new examples of topics and correct paper classifications where wrong. In this way the training set improves over time.

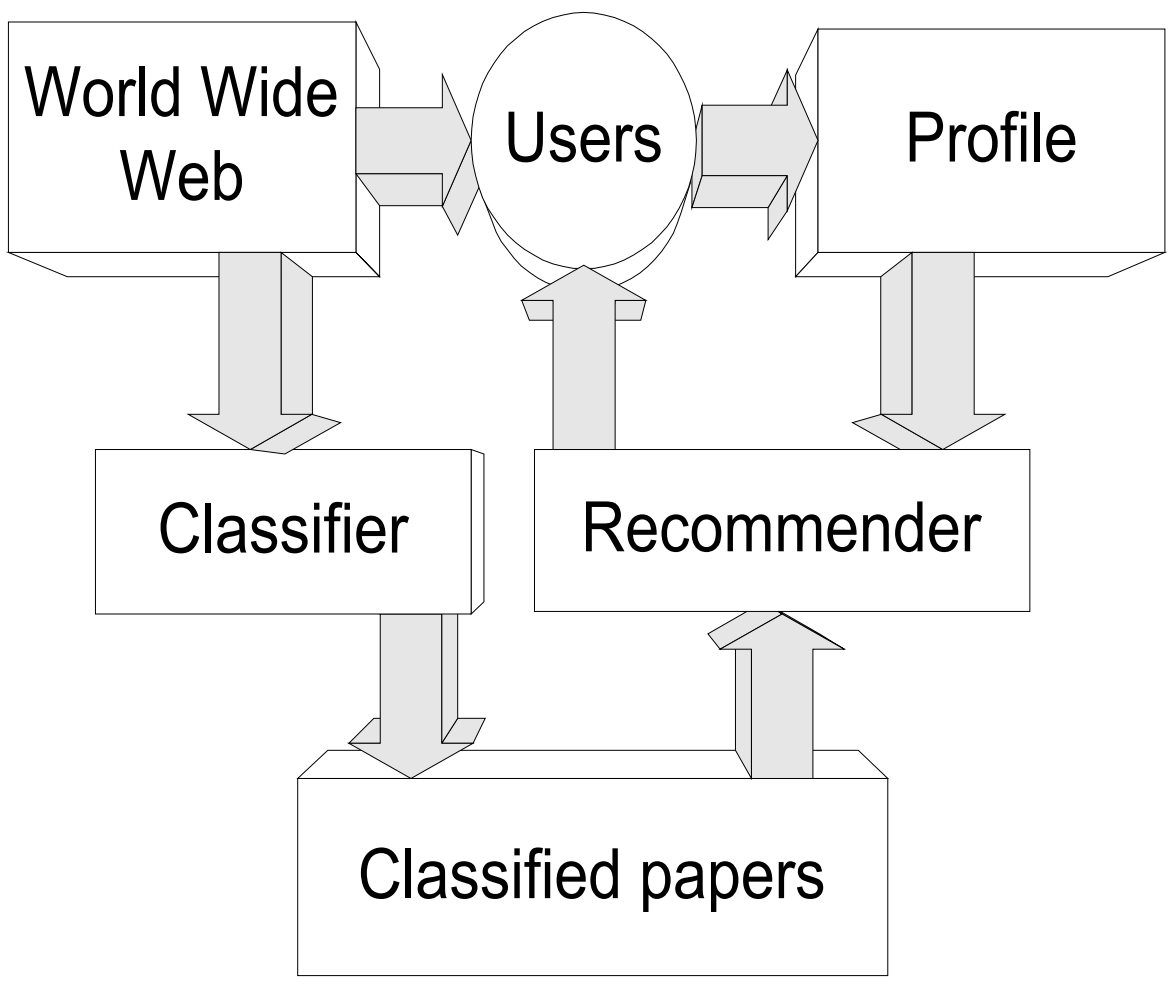

Figure 1 The Quickstep system

### Empirical Evaluation

The current literature lacks many clear results as to the extent knowledge-based approaches assist real-world systems, where noisy data and differing user opinions exist. For this reason we decided to compare the use of an ontology against a simple flat list, to provide some empirical evidence as to the effectiveness of this approach.

Two experiments are detailed within this paper. The first has 14 subjects, all using the Quickstep system for a period of 1.5 months. The second has 24 subjects, again over a period of 1.5 months.

Both experiments divide the subjects into two groups.

The first group uses a flat, extensible list of paper topics. Any new examples, added via explicit feedback, use this flat list to select from. The users are free to add to the list as needed.

The second group uses a fixed size topic ontology (based on the dmoz open directory project hierarchy [6]). Topics are selected from a hierarchical list based on the ontology. Interest profiles of this group take into account the super classes of any browsed topics.

Performance metrics are measured over the duration of the trial, and thus the effectiveness of both groups compared.

## APPROACH

### The Quickstep System

Quickstep is a hybrid recommendation system, combining both content-based and collaborative filtering techniques. Since both web pages and user interests are dynamic in nature, catalogues, rule-bases and static user profiles would quickly become out of date. A recommender system approach thus appeared well suited to our problem.

Explicit feedback on browsed papers would be too intrusive, so unobtrusive monitoring is used providing positive examples of pages the user typically browses. Many users will be using the system at once, so it is sensible to share user interest feedback and maintain a common pool of labelled example papers (provided by the users as examples of particular paper topics).

Since there are positive examples of the kind of papers users are interested in, we have a labelled training set. This is ideal for supervised learning techniques, which require each training example to have a label (the labels are then used as classification classes). The alternative, unsupervised learning, is inherently less accurate since it must compute likely labels before classification (e.g. clustering techniques). We shall use a term vector representation, common in machine learning, to represent a research paper. A term vector is a list of word weights, derived from the frequency that the word appears within the paper.

We could have used a binary classification approach, with classes for "interesting" and "not interesting". This would have led to profiles consisting of two term vectors, one representing the kind of thing the user is interested in (computed from the positive examples) and the other what the user is not interested in (computed from the negative examples). Recommendations would be those page vectors that are most similar to the interesting class vector. The binary case is the simplest class representation, and consequently produces the best classification results when compared with multi-class methods.

One problem with such a representation is that the explicit knowledge of which topics the user is interested in is lost, making it hard to benefit from any prior knowledge we may know about the domain (such as the paper topics). With Quickstep, we have chosen a multi-class representation, with each class representing a research paper topic. This allows profiles that consist of a human understandable list of topics. The classifier assigns each paper a class based on which class vector it is most similar to. Recommendations are selected from papers classified as belonging to a topic of interest.

The profile itself is computed from the correlation between browsed papers and paper topics. This correlation leads to a topic interest history, and a simple time-decay function allows current topics to be computed.

### Details of Specific Techniques Used

#### *Research Paper Representation*

Research papers are represented as term vectors, with term frequency / total number of terms used for a terms weight. To reduce the dimensionality of the vectors, frequencies less than 2 are removed, standard Porter stemming [18] applied to remove word suffixes and the SMART [22] stop list used to remove common words such as "the". These measures are commonly used in information systems; van Rijsbergen [24] and Harman [9] provide a good discussion of these issues.

Vectors with 10-15,000 terms were used in the trials along with training set sizes of about 200 vectors. Had we needed more dimensionality reduction, the popular term frequency-inverse document frequency (TF-IDF) weighting could be used (term weights below a threshold being removed) or latent semantic indexing (LSI).

Only Postscript and PDF formats (and compressed formats) are supported, to avoid noisy HTML pages. This makes classification easier, at the expense of HTML only papers.

#### *Research Paper Classification*

The classification requirements are for a multi-class learning algorithm learning from a multi-labelled training set. To learn from a training set, inductive learning is required. There are quite a few inductive learning techniques to choose from, including information theoretic ones (e.g. Rocchio classifier), neural networks (e.g. backpropagation), instance-based methods (e.g. nearest neighbour), rule learners (e.g. RIPPER), decision trees (e.g. C4.5) and probabilistic classifiers (e.g. naive Bayes).

Multiple classifier techniques such as boosting [7] exist as well, and have been shown to enhance the performance of individual classifiers.

After reviewing and testing many of the above options, we decided to use a nearest neighbour technique. The nearest neighbour approach is well suited to our problem, since the training set must grow over time and consists of multi-class examples. Nearest neighbour algorithms also degrade well, with the next closest match being reported if the correct one is not found. The IBk algorithm [1] we chose outperformed naive Bayes and a J48 decision tree in our tests. We also use the boosting technique AdaBoostM1 [7], which works well for multi-class problems if the boosted classifier is strong enough. We found that boosting always improved the base classifiers performance in our tests.

Nearest neighbour algorithms represent instances of documents as term vectors within a term vector space. Proximity of vectors within this term vector space indicates similarity. To classify a new paper, the vector distance from each example instance is calculated, and the closest neighbours returned as the most likely classes. Inverse distance weighting is used to decrease the likelihood of choosing distant neighbours.

AdaBoostM1 extends AdaBoost to handle multi-class cases since AdaBoost itself is a binary classifier. AdaBoostM1 repeatedly runs a weak learning algorithm (in this case the IBk classifier) for a number of iterations over various parts of the training set. The classifiers produced (specialized for particular classes) are combined to form a single composite classifier at the end.

*Profiling Algorithm*

The profiling algorithm performs correlation between the paper topic classifications and user browsing logs. Whenever a research paper is browsed that has a classified topic, it accumulates an interest score for that topic. Explicit feedback on recommendations also accumulates interest values for topics. The current interest of a topic is computed using the inverse time weighting algorithm below, applied to the user feedback instances.

$$\text{Topic interest} = \sum_{1..\text{no of instances}}^{n} \text{Interest value}(n) \,/\, \text{days old}(n)$$

Interest values   Paper browsed = 1
Recommendation followed = 2
Topic rated interesting = 10
Topic rated not interesting = -10

The profile for each user consists of a list of topics and the current interest values computed for them (see below). The interest value weighting was chosen to provide sufficient weight for an explicit feedback instance to dominate for about a week, but after that browsed URL's would again become dominant. In this way, the profile will adapt to changing user interests as the trial progresses.

Profile = (<user>,<topic>,<topic interest value>)*

e.g.   ((someone,hypertext,-2.4)
(someone,agents,6.5)
(someone,machine learning,1.33))

If the user is using the ontology based set of topics, all super classes gain a share when a topic receives some interest. The immediate super class receives 50% the main topics value. The next super class receives 25% and so on until the most general topic in the is-a hierarchy is reached. In this way, general topics are included in the profile rather than just the most specific ones, producing a more rounded profile.

*Recommendation Algorithm*

Recommendations are formulated from a correlation between the users current topics of interest and papers classified as belonging to those topics. A paper is only recommended if it does not appear in the users browsed URL log, ensuring that recommendations have not been seen before. For each user, the top three interesting topics are selected with 10 recommendations made in total (making a 4/3/3 split of recommendations). Papers are ranked in order of the recommendation confidence before being presented to the user.

Recommendation confidence = classification confidence * topic interest value

The classification confidence is computed from the AdaBoostM1 algorithm's class probability value for that paper (somewhere between 0 and 1).

*Research Paper Topic Ontology*

The research paper topic ontology is based on the dmoz [6] taxonomy of computer science topics. It is an is-a hierarchy of paper topics, up to 4 levels deep (e.g. an "interface agents" paper is-a "agents" paper). Pre-trial interviews formed the basis of which additional topics would be required. An expert review by two domain experts validated the ontology for correctness before use in the trials.

*Feedback and the Quickstep Interface*

Recommendations are presented to the user via a browser web page. The web page applet loads the current recommendation set and records any feedback the user provides. Research papers can be jumped to, opening a new browser window to display the paper URL. If the user likes/dislikes the paper topic, the interest feedback combo-box allows "interested" or "not interested" to replace the default "no comment". Finally, the topic of each paper can be changed by clicking on the topic and selecting a new one from a popup menu. The ontology group has a hierarchical popup menu; the flat list group has a single level popup menu.

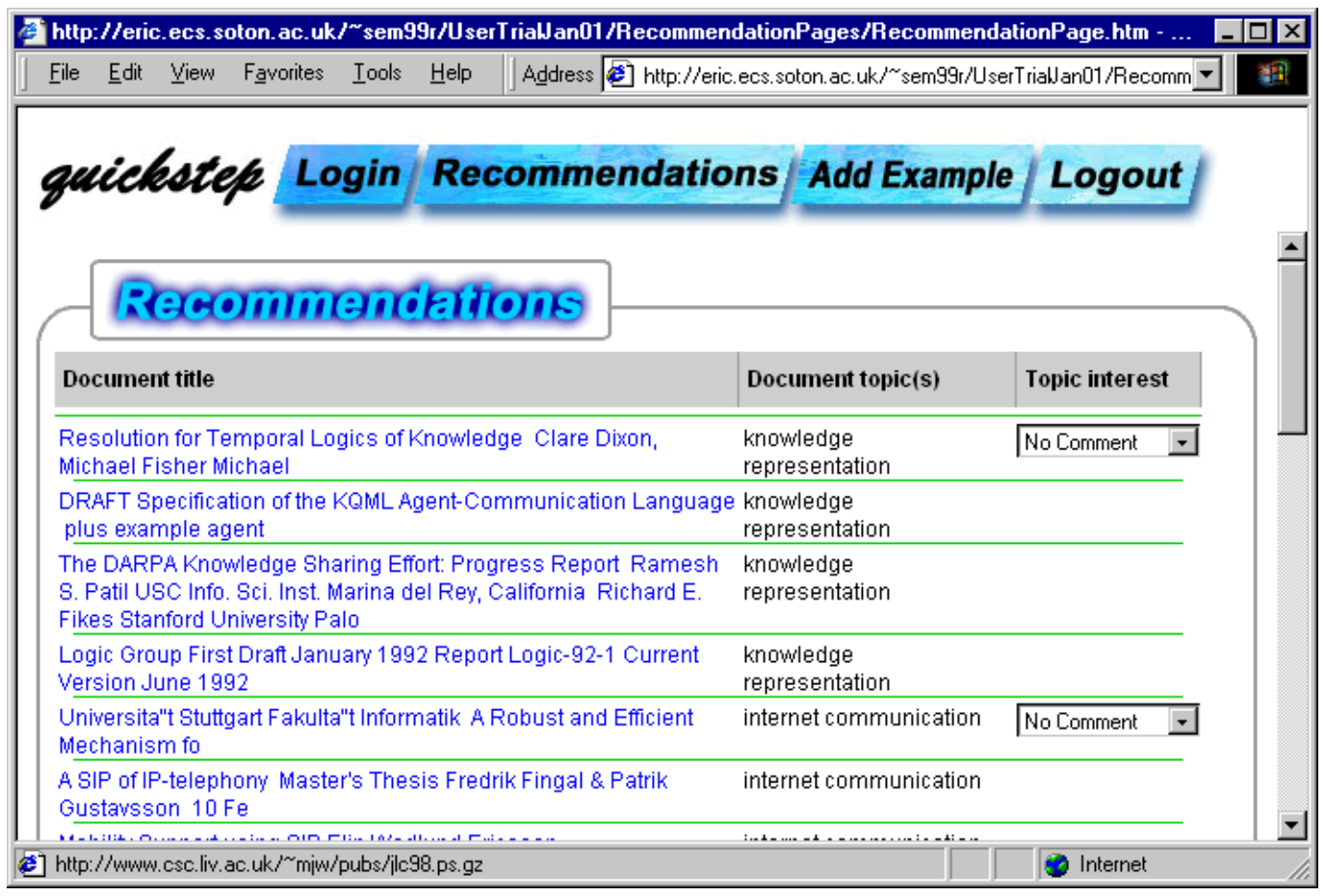

Figure 2 Quickstep's web-based interface

New examples can be added via the interface, with users providing a paper URL and a topic label. These are added to the groups training set, allowing users to teach the system new topics or improve classification of old ones.

All feedback is stored in log files, ready for the profile builders run. The feedback logs are also used as the primary metric for evaluation. Interest feedback, topic corrections and jumps to recommended papers are all recorded.

## EVALUATION

### Details of the Two Trials

Two trials were conducted to assess empirically both the overall effectiveness of the Quickstep recommender system and to quantify the effect made by use of the ontology.

The first trial used 14 subjects, consisting of researchers from the IAM research laboratory. A mixture of 2nd year postgraduates up to professors was taken, all using the Quickstep system for a duration of 1.5 months.

The second trial used 24 subjects, 14 from the first trial and 10 more 1st year postgraduates, and lasted for 1.5 months. Some minor interface improvements were made to make the feedback options less confusing.

The pre-trial interview obtained details from subjects such as area of interest and expected frequency of browser use.

The purpose of the two trials was to compare a group of users using an ontology labelling strategy with a group of users using a flat list labelling strategy. Subject selection for the two groups balanced the groups as much as possible, evening out topics of interest, browser use and research experience (in that order of importance). Both groups had the same number of subjects in them (7 each for the pilot trial, 12 each for the main trial).

In the first trial, a bootstrap of 103 example papers covering 17 topics was used. The bootstrap examples were obtained from bookmarks requested during the pre-trial interview.

In the second trial, a bootstrap of 135 example papers covering 23 topics was used. The bootstrap training set was updated to include examples from the final training sets of the first trial. The first trials classified papers were also kept, allowing a bigger initial collection of papers from which to recommend in the second trial.

Both groups had their own separate training set of examples, which diverged in content as the trial progressed. The classifier was run twice for each research paper, classifying once with the flat list groups training set and once with the ontology groups training set. The classifier algorithm was identical for both groups; only the training set changed.

The system interface used by both groups was identical, except for the popup menu for choosing paper topics. The ontology group had a hierarchical menu (using the ontology); the flat list group had a single layer menu.

The system recorded the times the user declared an interest in a topic (by selecting "interesting" or "not interesting"), jumps to recommended papers and corrections to the topics of recommended papers. These feedback events were date stamped and recorded in a log file for later analysis, along with a log of all recommendations made. Feedback recording was performed automatically by the system, whenever the subjects looked at their recommendations.

### Experimental Data

Since feedback only occurs when subjects check their recommendations, the data collected occurs at irregular dates over the duration of the trial. Cumulative frequency of feedback events is computed over the period of the trial, allowing trends to be seen as they develop during the trial. Since the total number of jumps and topics differ between the two groups, the figures presented are normalized by dividing by the number of topics (or recommendations) up to that date. This avoids bias towards the group that provided feedback most frequently.

Figure 3 shows the topic interest feedback results. Topic interest feedback is where the user comments on a recommended topic, declaring it "interesting" or "not interesting". If no feedback is offered, the result is "no comment".

Topic interest feedback is an indication of the accuracy of the current profile. When a recommended topic is correct for a period of time, the user will tend to become content with it and stop rating it as "interesting". On the other hand, an uninteresting topic is likely to always attract a "not interesting" rating. Good topics are defined as either "no comment" or "interesting" topics. The cumulative frequency figures are presented as a ratio of the total number of topics recommended. The not interesting ratio (bad topics) can be computed from these figures by subtracting the good topic values from 1.

The ontology groups have a 7 and 15% higher topic acceptance. In addition to this trend, the first trial ratios are about 10% lower than the second trial ratios.

Figure 4 shows the jump feedback results. Jump feedback is where the user jumps to a recommended paper by opening it via the web browser. Jumps are correlated with topic interest feedback, so a good jump is a jump to a paper on a good topic. Jump feedback is an indication of the quality of the recommendations being made as well as the accuracy of the profile. The cumulative frequency figures are presented as a ratio of the total number of recommendations made.

There is a small 1% improvement in good jumps by the ontology group. Both trials show between 8-10% of recommendations leading to good jumps.

Figure 5 shows the topic correction results. Topic corrections are where the user corrects the topic of a recommended paper by providing a new one. A topic correction will add to or modify a groups training set so that the classification for that group will improve. The number of corrections made is an indication of classifier accuracy. The cumulative frequency figures are presented as a ratio of the total number of recommended papers seen.

Although the flat list group has more corrections, the difference is only by about 1%. A clearer trend is for the flat list group corrections to peak around 10-20 days into the trial, and for both groups to improve as time goes on.

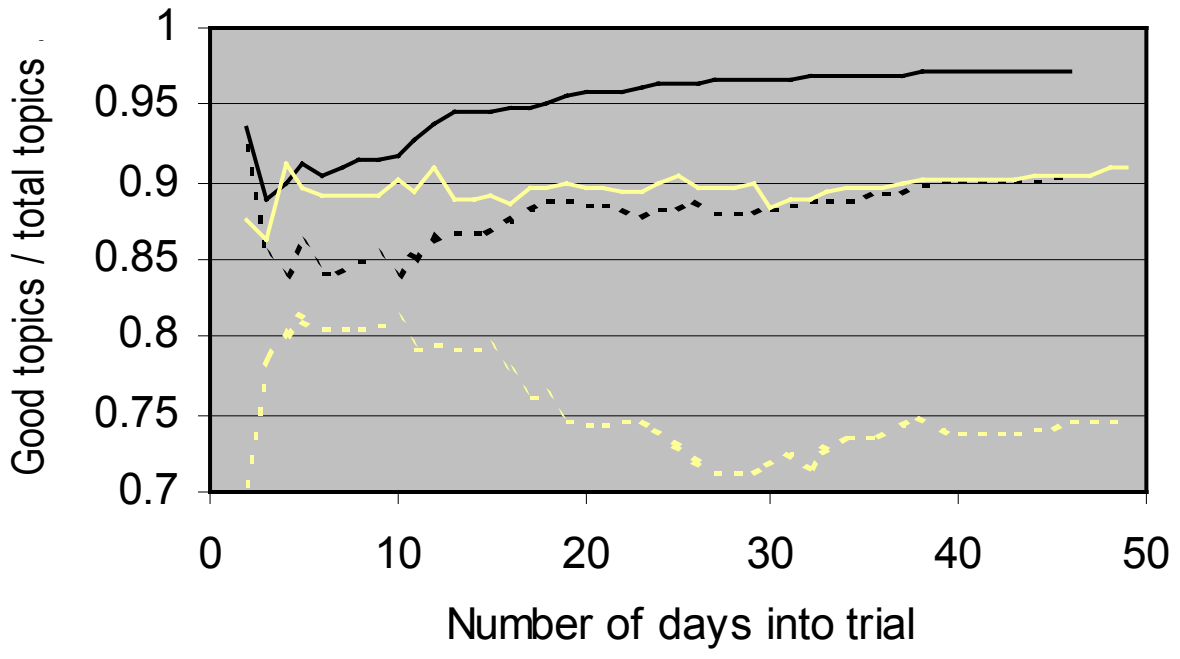

Figure 3 Ratio of good topics / total topics

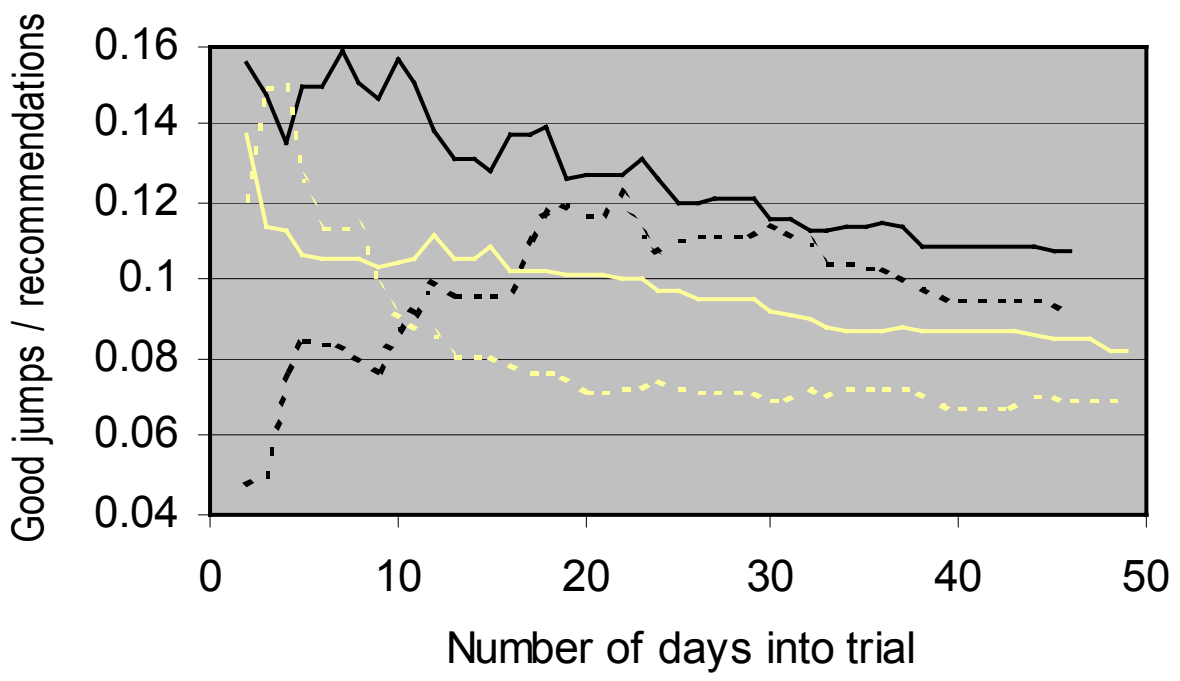

Figure 4 Ratio of good jumps / total recommendations

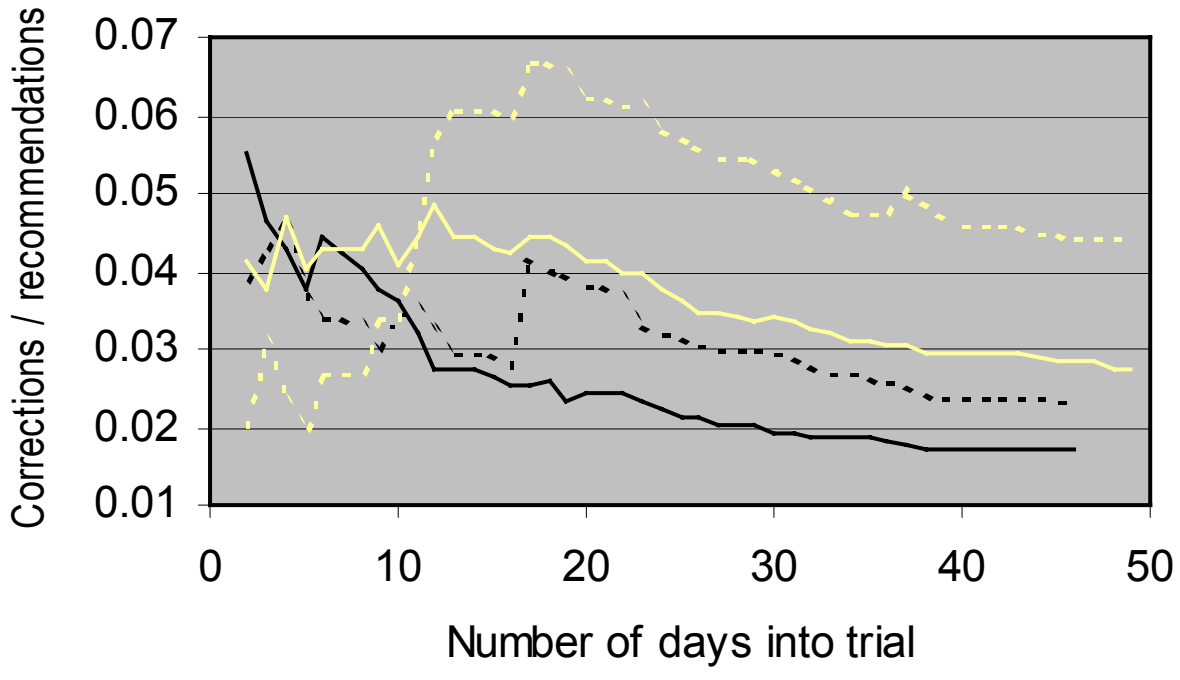

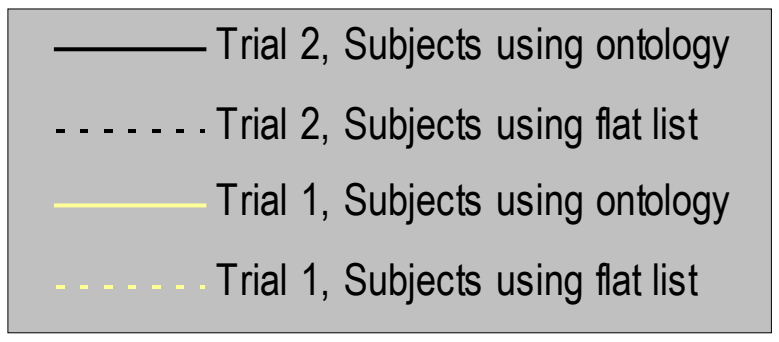

Figure 5 Ratio of topic corrections / total recommendations

A cross-validation test was run on each group's final training set, to assess the precision and recall of the classifier using those training sets. The results are shown in table 1.

| *Group (trial)* | *Precision* | *Recall* | *Classes* |
|---|---|---|---|
| Trial 1, Ontology | 0.484 | 0.903 | 27 |
| Trial 1, Flat list | 0.52 | 1.0 | 25 |
| Trial 2, Ontology | 0.457 | 0.888 | 32 |
| Trial 2, Flat list | 0.456 | 0.972 | 32 |

Table 1 Classifier recall and precision upon trial completion

**Discussion of Trends Seen in the Experimental Data**

From the experimental data of both trials, several suggestive trends are apparent. The initial ratios of good topics were lower than the final ratios, reflecting the time it takes for enough log information to be accumulated to let the profile settle down. The ontology users were 7-15% happier overall with the topics suggested to them.

Our hypothesis for the ontology group's apparently superior performance is that the is-a hierarchy produces a rounder, more complete profile by including general super class topics when a specific topic is browsed by a user. This in turn helps the profiler to discover a broad range of interests, rather than just latching onto one correct topic.

The first trial showed fewer good topics than the second trial (about a 10% difference seen by both groups). We think this is because of interface improvements made for the second trial, where the topic feedback interface was made less confusing. Subjects were sometimes rating interesting topics as not interesting if the paper quality was poor. As there are more poor quality papers than good quality ones, this introduced a bias to not interesting topic feedback resulting in a lower overall ratio.

About 10% of recommendations led to good jumps. Since 10 recommendations were given to the users at a time, on average one good jump was made from each set of recommendations received. As with the topic feedback, the ontology group again was marginally superior but only by a 1% margin. We think this smaller difference is due to people having time to follow only 1 or 2 recommendations. Thus, although the ontology group has more good topics, only the top topic of the three recommended will really be looked at; the result is a smaller difference between the good jumps made and the good topics seen.

The flat list group has a poor correction / recommendation ratio 10-20 days into the trial. We think this is due to new topics being added to the system. Most new topics were added after the users became familiar with the system, and know which topics they feel are missing. The familiarization process appeared to take about 10 days. The classification accuracy of these new topics is poor until enough examples have been entered, typically after another 10 days.

The ontology group has about 1% fewer corrections for both trials. This is small difference may indicate the utility of imposing a uniform conceptual model of paper topics on the subjects (by using the common topic hierarchy). Classifying papers is a subjective process, and will surely be helped if people have similar ideas as to where topics fit in a groups overall classification scheme.

These preliminary results need to be extended so as to enable the application of more rigorous statistical analysis. Nevertheless, we believe the trend in the data to be encouraging as to the utility of ontologies in recommender systems.

When compared with other published systems, the classification accuracy figures are similar, if on the low side (primarily because we use multi-class classification). Nearest neighbour systems such as NewsDude [3] and Personal Webwatcher [14] report 60-90% classification accuracy based on binary classification. The higher figures tend to be seen with benchmark document collections, not real-world data. NewsWeeder [12] reports 40-60% classification accuracy using real user browsing data from two users over a period of time, so this would be the best comparison. If the number of classes we classify is taken into consideration, our system compares well.

Multi-class classification is not normally applied to recommender systems making direct comparison of similar systems difficult. We would have liked to compare the usefulness of our recommender to that of other systems, but the lack of published experimental data of this kind means we can only usefully compare classification accuracy.

## CONCLUSIONS

Most recommender systems use a simple binary class approach, using a user profile of what is interesting or not interesting to the user. The Quickstep recommender system uses a multi-class approach, allowing a profile in terms of domain concepts (research paper topics) to be built. The multi-class classification is less accurate than other binary classification systems, but allows class specific feedback and the use of domain knowledge (via an is-a hierarchy) to enhance the profiling process.

Two experiments are performed in a real work setting, using 14 and 24 subjects over a period of 1.5 months. The results suggest how using an ontology in the profiling process results in superior performance over using a flat list of topics. The ontology users tended to have more "rounder" profiles, including more general topics of interest that were not directly suggested. This increased the accuracy of the profiles, and hence usefulness of the recommendations.

The overall performance compares reasonably with other recommender systems.

### Related Work

Collaborative recommender systems utilize user ratings to recommend items liked by similar people. Examples of collaborative filtering are PHOAKS [23], which recommends web links mentioned in newsgroups and Group Lens [11], which recommends newsgroup articles.

Content-based recommender systems recommend items with similar content to things the user has liked before. Examples of content-based recommendation are Fab [2], which recommends web pages and ELFI [19], which recommends funding information from a database.

Personal web-based agents such as Letizia [13], Syskill & Webert [17] and Personal Webwatcher [14] track the users browsing and formulate user profiles. Profiles are constructed from positive and negative examples of interest, obtained from explicit feedback or heuristics analysing browsing behaviour. They then suggest which links are worth following from the current web page by recommending page links most similar to the users profile.

News filtering agents such as NewsWeeder [12] and News Dude [3] recommend news stories based on content similarity to previously rated examples.

Systems such as CiteSeer [4] use content-based similarity matching to help search for interesting research papers within a digital library. Ontologies are also used to improve content-based search, as seen in OntoSeek [8].

Mladenic [15] provides a good survey of text-learning and agent systems, including content-based and collaborative approaches.

### Future Direction of Work

The next step for this work is to run more trials and perform rigorous statistical analysis on the results. As the subjects increase in number, we can become increasingly confident of the power of the effects we are seeing.

Paper quality ratings will be elicited from users, so once an interesting topic has been discovered, good quality papers can be recommended before poorer quality papers.

The idea of building a profile that is understandable by the users could be extended to actually visualizing the knowledge contained within it. This will allow the recommender to engage the user in a dialogue about what exactly they are interested in. The knowledge elicited from this dialogue should allow further improvements to the recommendations made. Additionally, visualizing the profile knowledge will allow users to build a better conceptual model of the system, helping to engender a feeling of control and eventually trust in the system.

## ACKNOWLEDGEMENTS

This work is funded by EPSRC studentship award number 99308831.